\documentclass[conference,times,mathptm,psfig]{IEEEtran}
\IEEEoverridecommandlockouts
\usepackage{cite}
\usepackage{amsmath,amssymb,amsfonts}
\usepackage{algorithmic}
\usepackage{graphicx}
\usepackage{textcomp}
\usepackage{xcolor}
\usepackage{multirow}
\def\BibTeX{{\rm B\kern-.05em{\sc i\kern-.025em b}\kern-.08em
    T\kern-.1667em\lower.7ex\hbox{E}\kern-.125emX}}
\begin{document}

\title{Impact of Physical Activity on Quality of Life During Pregnancy: A Causal ML Approach
\\
}

\author{\IEEEauthorblockN{Kianoosh Kazemi$^{1}$, Iina Ryhtä$^{2}$, Iman Azimi$^{3}$}, Hannakaisa Niela-Vil\'en$^{2}$, Anna Axelin$^{2}$, Amir M. Rahmani$^{3}$, Pasi Liljeberg$^{1}$\\
$^{1}$Department of Computing, University of Turku, Finland, \\
$^{2}$Department of Nursing Science, University of Turku, Turku, Finland\\
$^{3}$Department of Computer Science, University of California, Irvine, USA\\
\{kianoosh.k.kazemi, iikrry, hmniel, anmaax, pasi.liljeberg\}@utu.fi, \{azimii, a.rahmani\}@uci.edu

% \IEEEauthorblockA{\textit{dept. of Computing} \\
% \textit{University of Turku}\\
% Turku, Finland \\
% kianoosh.k.kazemi@utu.fi}
% \and
% \IEEEauthorblockN{Iina Ryhtä}
% \IEEEauthorblockA{\textit{dept. of Nursing} \\
% \textit{University of Turku}\\
% Turku, Finland \\
% \and
% \IEEEauthorblockN{Iman Azimi}
% \IEEEauthorblockA{\textit{dept. of Computer Science} \\
% \textit{University of California}\\
% Irvine, USA \\
% \and
% \IEEEauthorblockN{Hannakaisa Niela-vilen}
% \IEEEauthorblockA{\textit{dept. of Nursing} \\
% \textit{University of Turku}\\
% Turku, Finland \\
% \and
% \IEEEauthorblockN{Anna Axelin}
% \IEEEauthorblockA{\textit{dept. of Nursing} \\
% \textit{University of Turku}\\
% Turku, Finland \\
% \and
% \IEEEauthorblockN{Amir M. Rahmani}
% \IEEEauthorblockA{\textit{dept. of Computer Science} \\
% \textit{University of California}\\
% Irvine, USA \\
% \and
% \IEEEauthorblockN{Pasi Liljeberg}
% \IEEEauthorblockA{\textit{dept. of Computing} \\
% \textit{University of Turku}\\
% Turku, Finland \\
}

\maketitle

\begin{abstract}
The concept of Quality of Life (QoL) refers to a holistic measurement of an individual's well-being, incorporating psychological and social aspects. Pregnant women, especially those with obesity and stress, often experience lower QoL. Physical activity (PA) has shown the potential to enhance the QoL. However, pregnant women who are overweight and obese rarely meet the recommended level of PA. Studies have investigated the relationship between PA and QoL during pregnancy using correlation-based approaches. These methods aim to discover spurious correlations between variables rather than causal relationships. Besides, the existing methods mainly rely on physical activity parameters and neglect the use of different factors such as maternal (medical) history and context data, leading to biased estimates.  Furthermore, the estimations lack an understanding of mediators and counterfactual scenarios that might affect them. In this paper, we investigate the causal relationship between being physically active (treatment variable) and the QoL (outcome) during pregnancy and postpartum. To estimate the causal effect, we develop a Causal Machine Learning method, integrating causal discovery and causal inference components. The data for our investigation is derived from a long-term wearable-based health monitoring study focusing on overweight and obese pregnant women. The causal graph is generated using the causal discovery method and modified with the help of a domain expert team to accommodate the mediators in the causal model. The machine learning (meta-learner) estimation technique is used to estimate the causal effect. Our result shows that performing adequate physical activity during pregnancy and postpartum improves the QoL by units of 7.3 and 3.4 on average in physical health and psychological domains, respectively. In the final step, four refutation analysis techniques are employed to validate our estimation.

\end{abstract}


\section{Introduction}

Quality of Life (QoL) is a comprehensive concept that provides a holistic evaluation of an individual's well-being, emphasizing the biopsychosocial aspects of health \cite{b1}. As defined by the World Health Organization (WHO), QoL reflects an individual's perspective on their life, taking into account personal goals and expectations. Research indicates that pregnant women generally experience lower QoL compared to population in general \cite{b2}.

Factors such as obesity, stress, anxiety, and sleep disturbances can negatively impact on QoL during pregnancy, while engaging in  regular physical activity (PA) can enhance it \cite{b2}. The significance of improving the QoL of pregnant women lies in its potential to influence the well-being not only of the mother but also of the entire family. It is recommended that pregnant women aim to achieve 150 minutes of moderate PA per week. However, it has been studied that \cite{b3} pregnant women with overweight and obesity met the PA recommendations rarely. In this regard, enhancing the PA with overweight and obese pregnant women is important since PA is known to have many short- and long-term benefits for health, as well as for more holistic outcomes, e.g., QoL.

Several studies have been conducted to examine how PA impacts pregnant women's quality of life. For example, descriptive statistics measures, such as mean value, standard deviation, confidence intervals, and standardized mean differences, were used to explore the associations between PA and the QoL \cite{b4, b5,b6,b7}. Additionally, Pearson correlation coefficients have been utilized to explore the association between physical activity and QoL \cite{b8}. In this study, the Pregnancy Physical Activity Questionnaire tool was used to assess and measure the physical activity levels of 141 pregnant women. In another study \cite{b9}, the relationship between PA and the QoL of pregnant women was assessed by employing nonparametric tests such as one-way analysis of variance (ANOVA), Chi-square test, and linear regression model.

There are several significant challenges involved in relying solely on correlation analysis when analyzing a complex phenomenon with multiple causal pathways. The causal relationships between variables must be disentangled rather than being reliant on spurious correlations. In other words, the conventional methods for estimating the effect of PA on QoL during pregnancy rely mainly on correlation between variables and are inadequate to assess the true underlying values. Moreover, most of the conventional techniques neglect the use of different factors such as maternal (medical) history and context data, including background information. Subsequently, these techniques generate biased estimates. Many studies \cite{b10} have explored the limitations of these methods, pointing out the need to go beyond just identifying data associations and understanding the processes that generate this data and the causal relationships between variables. Statistical approaches have their limitations in revealing the direct and indirect causes linking pregnant women's QoL, PA, and other variables and in identifying and adjusting for potential biases.

Furthermore, the existing studies relying on self-report questionnaires \cite{b6, b11}  face limitations, including subjective reporting influenced by recall and social desirability bias, as well as limited precision in responses. Particularly in assessing pregnant women's PA, subjective methods like interviews and self-report questionnaires may inaccurately reflect PA. To address these challenges, there is a growing trend toward the use of various eHealth monitoring devices \cite{b12}. Wearables, measuring physical and health conditions objectively \cite{b13, b14}, provide continuous and accurate data, overcoming the drawbacks of traditional subjective methods. This shift enhances the quality of data, reduces dropout rates, and contributes to better representation of health-related data on the levels and changes of activity during pregnancy and the postpartum period. We believe that a Causal Machine Learning (CML) method is required to exploit  multi-modal data collection, such as wearable-derived and subjective data (i.e., maternal (medical) history and context data), while minimizing the bias of estimates.

In this paper, we examine the causal relationship between QoL and the PA of pregnant women who are overweight and obese. To collect the required data, a long-term health monitoring of 48 pregnant women was conducted. In this study, we develop a CML approach consisting of two main components: causal discovery and causal inference. In the first step, the causal discovery is exploited to recover the underlying causal relationship between PA (the treatment variable) and QoL (the outcome variable) utilizing the construction of causal graphs. Then, we use the knowledge of the nursing expert team to modify the constructed causal graph that clearly defines the set of variables pertinent to both the PA and QoL. In the next step, the estimation algorithm (i.e., as a part of the causal inference) is utilized to calculate the average treatment effects for our dataset. By applying the CML method, the causal impact of PA on the QoL during pregnancy and post-partum in this collected dataset is examined. Our hypotheses are further validated through a variety of refutation techniques, allowing us to draw well-informed conclusions.
\section{Method}
In this section, we first present our data collection methodology, followed by an explanation of the CML method. This study is part of a large intervention study where lifestyle intervention was developed and tested with overweight pregnant women \cite{b15}. In this paper, our focus is on the physical health and psychological health domains of QoL.

\subsection{ Data Collection}\label{II-A}
A long-term health monitoring study of overweight and obese pregnant women was used for this study. The aim is to examine the relationship between PA and QoL. Tracking vital signs, PA, and sleep was done using the OURA ring while participants engaged in their normal daily routines.

Recruitment started in April 2021 and ended in May 2023. Pregnant women with overweight and obesity (BMI$\mathrm{>}$25)  were recruited before pregnancy week 15. After explaining the study details, researchers provided OURA rings to participants with usage instructions. Data collection occurred through an online platform that we developed through which researchers could send questionnaires and surveys to participants at 15 gestational weeks, 34 gestational weeks, and 12 weeks post-delivery, along with continuous OURA ring monitoring from the recruitment until three months postpartum. Data from 48 pregnant women were used in our analysis. The participants' backgrounds are summarized in Table \ref{tab1}.

Data collection was conducted using the OURA ring \cite{b17}, a commercially available wearable ring designed for monitoring a range of health parameters and generating personalized daily health scores. The recorded daily data were transmitted to smartphones and web applications through a Bluetooth connection. In addition, women answered QoL questionnaires. QoL is measured by the World Health Organization Quality of Life (WHOQOL–BREF) questionnaire \cite{b1}. WHOQOL-BREF is a 26-item assessment tool categorized into four domains: physical health, psychological health, social relationships, and environment, scoring between 0-100; higher scores indicate higher QoL.  Moreover,  a questionnaire measuring depressive symptoms (EPDS) \cite{b18} was also completed at each timepoint via the developed platform. 

\begin{table}[tbp]
\caption{background information}
\begin{tabular}{|c|c|c|c|}
\hline
Characteristic                                                                                                                               & \begin{tabular}[c]{@{}c@{}}All participants \\ (n=48)\end{tabular}               & \begin{tabular}[c]{@{}c@{}}Active group\\  (n=22)\end{tabular}                  & \begin{tabular}[c]{@{}c@{}}Low active group\\  (n=26)\end{tabular}                \\ \hline
Age, Mean (SD)                                                                                                                               & 29.3 (4.6)                                                                       & 29.8 (5.2)                                                                      & 29.2 (4.0)                                                                        \\ \hline
\begin{tabular}[c]{@{}c@{}}BMI,  Mean (SD)\end{tabular}                                                                                    & 30.69 (4.74)                                                                     & 29.47 (4.51)                                                                    & 31.80 (4.76)                                                                      \\ \hline
\begin{tabular}[c]{@{}c@{}}Married or living \\ with a partner\\ – relationship status–  \\ (yes) n (\%),\\  Mean (SD)\end{tabular}          & 47 (97.9)                                                                        & 22 (100.0)                                                                      & 24 (96.0)                                                                         \\ \hline
\begin{tabular}[l]{l@{}c@{}}Employment status\\  n (\%) Mean (SD)\\   $\bullet$  Working \\   $\bullet$  Unemployed \\   $\bullet$  Student \\   $\bullet$  Other\end{tabular} & \begin{tabular}[c]{@{}c@{}} \\ \\  38 (79.2)\\ 3 (6.3)\\ 2 (4.2)\\ 5 (10.4)\end{tabular} & \begin{tabular}[c]{@{}c@{}}\\ \\ 21 (95.5)\\ 0 (0.0)\\ 1 (4.5)\\ 0 (0.0)\end{tabular} & \begin{tabular}[c]{@{}c@{}}\\ \\17 (68.0)\\ 3 (12.0)\\ 0 (0.0)\\ 5 (20.0)\end{tabular} \\ \hline
\begin{tabular}[c]{@{}c@{}}EPDS at recruitment,\\  Mean (SD)\end{tabular}                                                                    & \begin{tabular}[c]{@{}c@{}}\\ 6.1 (4.1)\end{tabular}                         & \begin{tabular}[c]{@{}c@{}}\\ 5.4 (3.5)\end{tabular}                        & \begin{tabular}[c]{@{}c@{}}\\ 6.6 (4.6)\end{tabular}                          \\ \hline
\begin{tabular}[c]{@{}c@{}}Number of Children, \\ Mean (SD)\end{tabular}                                                                     & 1.20 (1.6)                                                                       & 1.3 (1.5)                                                                       & 1.14 (1.6)                                                                        \\ \hline
\end{tabular}
\label{tab1}
\end{table}

We separated participants into two distinct, non-overlapping subgroups: the active group and the low active group. Utilizing the OURA ring, PA data (such as activity type, intensity, duration, step counts, and the date/time of activity) were collected daily. We used the medium-intensity activity parameter (i.e., minutes per day that participants have performed activities with an intensity level equivalent to walking) collected by OURA to calculate the weekly duration of medium-intensity activity for each participant. Subsequently, a threshold of 150 minutes per week, in alignment with the recommended duration of moderate PA for pregnant women \cite{b3}, was established. This classification resulted in 22 women being placed in the active group, with the remaining 26 pregnant women assigned to the low active group.
Fig. \ref{fig_activity} presents a visual representation of the weekly active time for the two groups. Fig. \ref{fig_activity} highlights that the active group exhibits a longer duration of medium activity compared to the low active group throughout the phases of pregnancy and postpartum. Furthermore, in the vicinity of the due date, the activity duration converges, reaching a comparable level for both groups.
In this study, the following physical activity attributes are used for the analysis of causal effect:

\begin{figure}[htbp]
\centerline{\includegraphics[height = 4.60cm, width=6.75cm]{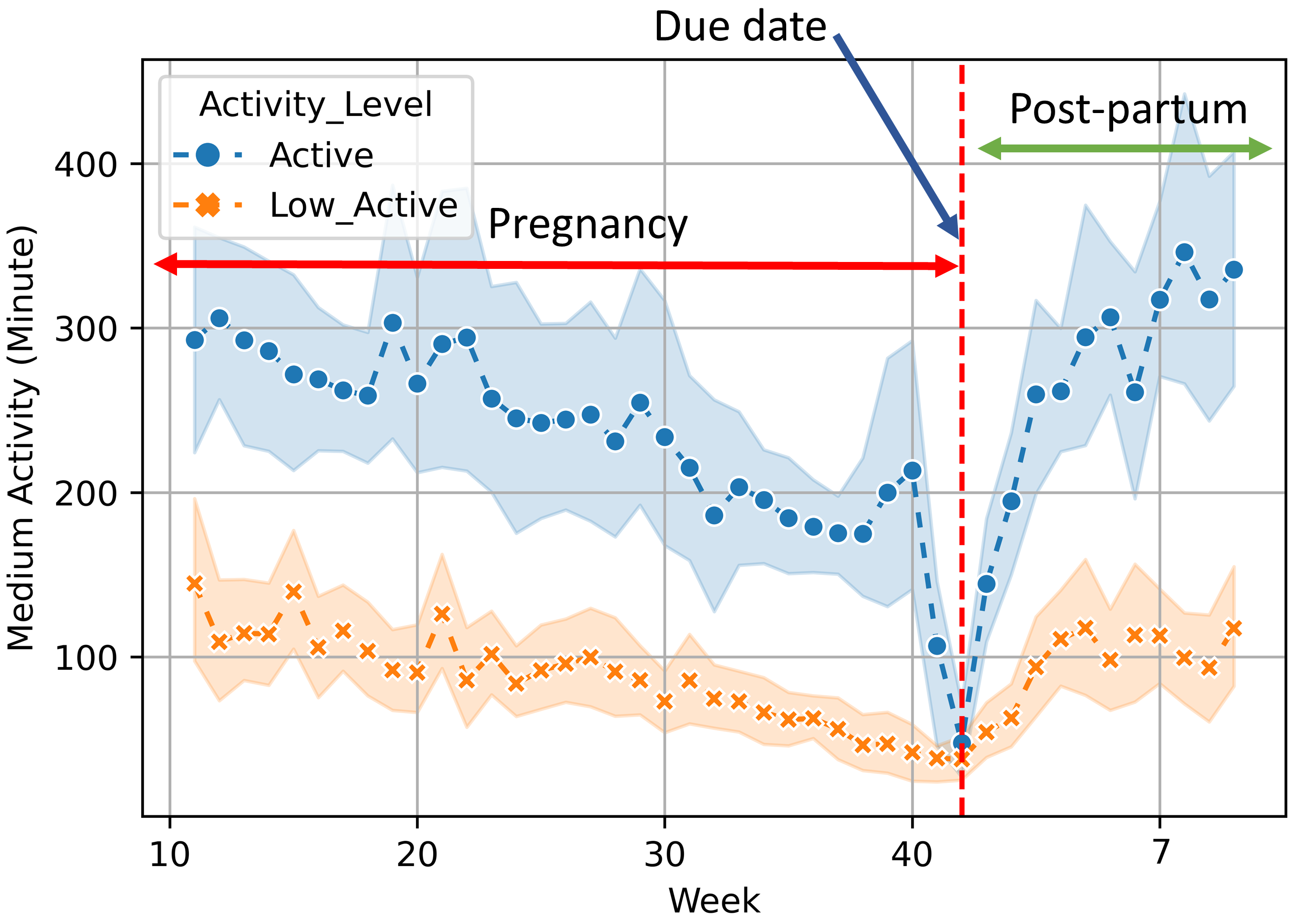}}
\caption{The activity duration for the Active vs. the Low active group.}
\label{fig_activity}
\end{figure}
\begin{itemize}
\item Step counts: Total number of steps taken by an individual per day.
\item Average-MET: Shows the daily average MET when the user is awake. METs (Metabolic Equivalents) are used to measure the intensity and energy expenditure of different physical activities.
\end{itemize}

This research followed the ethical principles outlined in the Declaration of Helsinki and the Finnish Medical Research Act (No 488/1999). It received collective approval from the Ethics Committee of the Hospital District of Southwest Finland and the maternity clinics (Protocol 113/1801/2020). Participants received comprehensive oral and written information prior to providing their written informed consent.

\subsection{Causal Discovery}
In this study, we aim to uncover the underlying causal relationships between QoL and PA during pregnancy by developing and applying causal discovery algorithms. These algorithms will enable us to produce concrete causal graphs that provide insight into how these relationships operate. Prior to employing the causal discovery algorithms, it is essential to normalize our data, ensuring its adherence to a standardized range. The objectives of the normalization process are three-fold, including enhancing uniformity of data, improving convergence during analysis and mitigating scale-related challenges. Furthermore, standardization is carried out to calculate the covariance matrix, which also serves as the basis for generating causal graphs.

In this study, we utilized three causal discovery algorithms: Peter-Clark (PC) \cite{b19, b20,b21}, Greedy Equivalence Search (GES) \cite{b21}, and Greedy Interventional Equivalence Search (GIES) \cite{b21}. These algorithms were selected based on the characteristics of the dataset and computational resources available to us. PC, a constrained-based causal discovery algorithm, initiates from a fully-connected graph and identifies potential direct causal relations among variables by employing conditional independence tests. Scalability and capability are two main features of the PCs algorithm, making this method suitable for working with large datasets. The GES and GIES algorithms build causal graphs from an empty graph, then add or remove edges based on a score, such as the Bayesian Information Criterion (BIC) \cite{b21}. GES algorithm can be used to analyze causal relationships of both linear and non-linear types with a wide range of variables. GIES \cite{b21}, a derivative of GES, is designed to reduce computational costs.

The causal graphs generated by the causal discovery component were analyzed and the paths were adjusted to identify strongly contributing mediators. PA affects QoL through a number of mediator variables. A team of nursing experts' input was utilized iteratively in this process.

\subsection{Causal Inference}
Causal inference attempts to estimate the causal effect of a treatment on a particular outcome while adjusting the effect for certain variables that could bias the estimate. In this study, a multi-mediator analysis \cite{b19} is utilized to estimate the causal impact of the treatment variable (physical activity) on the outcome (QoL score), taking into account the mediator variables identified in the causal inference component. Causal effects can be estimated using a variety of machine learning methods \cite{b19, b20,b21}. Our approach involves constructing a meta-learner method (i.e., T-learner, a machine-learning predictive model using different models for each treatment by splitting the data) for the multi-mediator analysis, including both the Average Treatment Effect (ATE) and mediator analysis. In ATE, average differences in outcomes are calculated between treatment and control groups to estimate the causal impact of a treatment variable \cite{b19, b20,b21}. An important technique in causal inference is mediating effects \cite{b19}, identifying how variables are influenced by each other. An analysis of the mediate effect investigates how two variables ($\mathrm{X}$  and $\mathrm{Y}$) interact through the mediating influence of a third variable ($\mathrm{Z}$). The mediator analysis includes three critical components to estimate the overall effect of $\mathrm{X}$  on $\mathrm{Y}$: 1) the Natural Direct Effect (NDE) of $X$ on $\mathrm{Y}$, 2) the Natural Indirect Effect (NIE) of $\mathrm{X}$  on $\mathrm{Y}$  via mediator $\mathrm{Z}$, and 3) the Total Effect (TE) defined by the structural causal model. TE consists of the same calculations as the ATE, including mediator calculations as well.

NDE is defined in (\ref{eq1}) as the expected change in $Y$ resulting from shifting $\mathrm{X}$ from $\mathrm{x}$ to $x^{\prime}$ while keeping the mediating variables at their previous values (i.e., $X = x$) before the shift from $\mathrm{x}$to $x^{\prime}$.
\begin{equation}
N D E_{x, x^{\prime}}(Y)=\sum\left[E\left(Y \mid x^{\prime}, z\right)-E(Y \mid x, z)\right] P(z \mid x) \label{eq1}
\end{equation}
When $\mathrm{X}$ is maintained constant at $X = x$, and $\mathrm{Z}$ (for each sample row) changes to the value it would have reached if $\mathrm{X}$ had been set to $\mathrm{X}$ = $x^{\prime}$, NIE is the expected change in $\mathrm{Y}$. Equation (\ref{eq2}) can be used to calculate this counterfactual concept. 

\begin{equation}
N I E_{x},{ }_{x^{\prime}}(Y)=\sum_{z} E(Y \mid x, z)\left[P\left(z \mid x^{\prime}\right)-P(z \mid x)\right]\label{eq2}
\end{equation}

Equation (\ref{eq3}) calculates mediators for the TE of $\mathrm{X}$ on $\mathrm{Y}$ in a Directed Acyclic Graph. Unlike linear systems, non-linear systems have no additive effect. To be more precise, TE represents the difference between direct and indirect effects (from $\mathrm{X}$ = $x^{\prime}$ to $\mathrm{X}$ = $\mathrm{x}$) of a reverse transition \cite{b19}. In summary, in the case of a transition between $\mathrm{X}$ = $\mathrm{x}$ and $\mathrm{X}$ = $x^{\prime}$ (e.g., two groups of active and low active), the TE measures the predicted change in $\mathrm{Y}$ due to the change in $\mathrm{X}$.

\begin{equation}
T E_{x, x^{\prime}}(Y)=N D E_{x, x^{\prime}}(Y)-N I E_{x, x^{\prime}}(Y)\label{eq3}
\end{equation}
\subsection{Refutation Analysis}
Our estimations are tested by refuting them against unverified assumptions to determine their sensitivity and robustness. As part of the refutation or counterargument approach, noise is added to the common cause variable or the treatment is replaced with a random variable in order to challenge the estimates. Performing this analysis helps identify potential biases or confounding factors that may affect the stability and robustness of the estimate. The p-value less than 0.05 indicates the failure of the refutation test, meaning the estimator and assumptions are invalid. Four refutation methods are used in our experiments \cite{b19}. Placebo Treatment: In this test, an independent random variable replaces the true treatment variable. It is expected that the new estimation goes toward zero; otherwise, the assumption is invalid. Data Subsets: This entails substituting the given dataset with a randomly selected subset. The expected result is that the new estimation remains unchanged; any deviation indicates an invalid assumption. Add Random Common Cause: An independent random variable as a common cause is added to the dataset. The new estimate is expected to remain unchanged; any changes signify an invalid assumption. Unobserved common cause: This approach introduces a supplemental dataset with a common cause between treatment and outcome. By adjusting values to match known impact strengths, the robustness is assessed via minimal estimator fluctuations. The expected result is that the new estimation remains unchanged. In addition, this refutation does not return a p-value.

\section{RESULTS}
This section presents the findings of our CML approach using different refutation techniques. To estimate causal effects using the proposed CML method, we employed four stages: 1) causal graph construction (output of causal discovery component), 2) the estimand identification using the causal graph, 3) causal impact computation (output of causal inference component), and 4) the estimate validation by manipulating assumptions (refutation analysis).

\begin{figure}[!tbp]
\centerline{\includegraphics[height = 5.60cm, width=6.25cm]{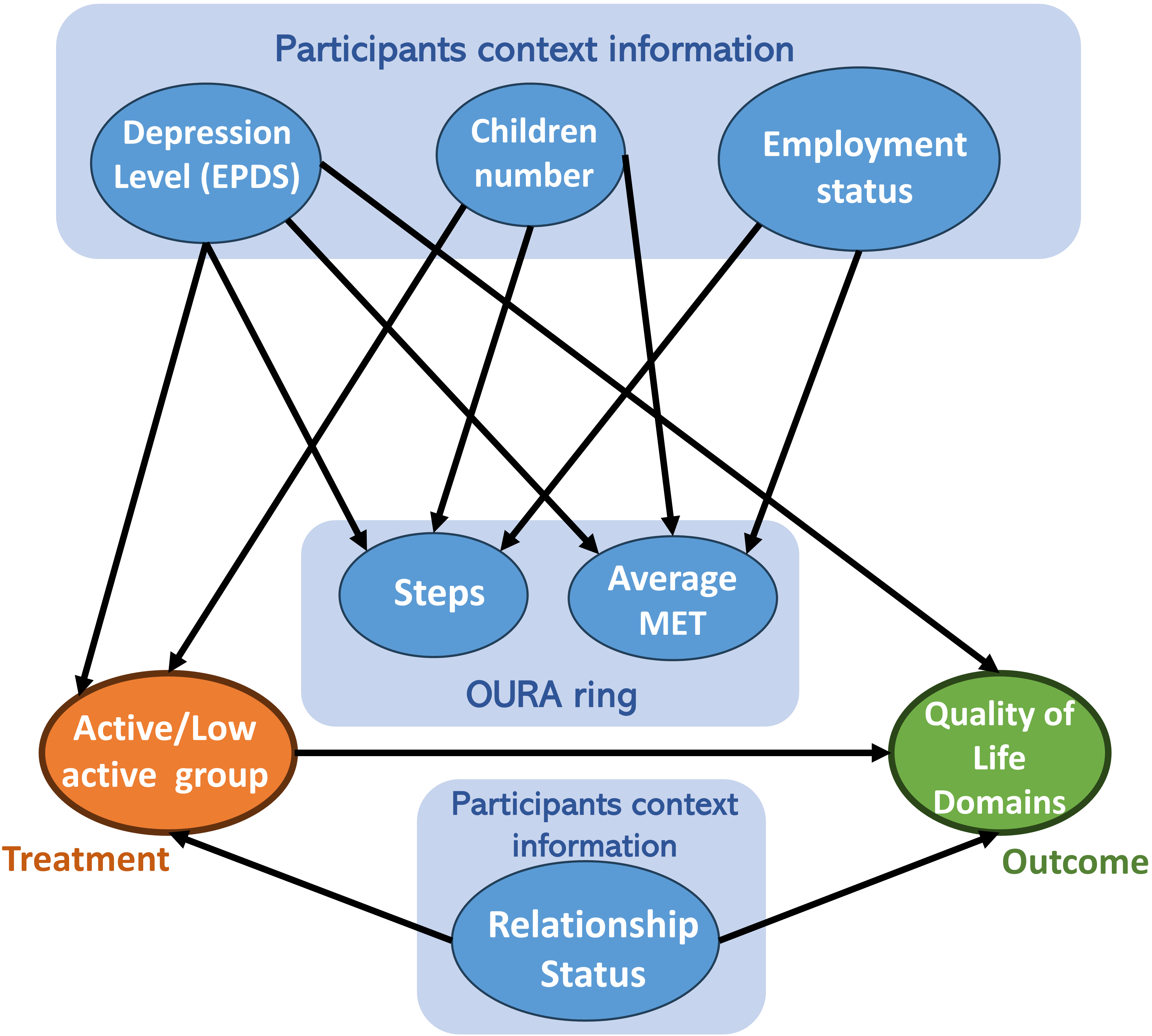}}
\caption{Causal graph.}
\label{fig}
\end{figure}

The objective of this study is to determine the causal relationship between PA (treatment variable) and QoL score (outcome) during pregnancy and post-partum, ranging from 0 to 100. To achieve this, we applied the CML method to the dataset (containing 48 pregnant women data) described in Section \ref{II-A}. We leveraged the causal discovery method to construct the causal graph. Then, by exploiting the knowledge of our nursing expert team, the causal graph was readjusted, as shown in Fig. \ref{fig}.

In the Causal graph (Fig. \ref{fig}), we included multiple mediators consisting of steps, average-met from the OURA ring dataset, and the number of children, work situation, and relationship status from the background information as well as depression level from the EPDS questionnaire. We computed the ATE using meta-learners (T-Learners) estimators. A light gradient boosting regressor (max depth = 2 and min child samples = 60) was used in the meta-learner to calculate the outcomes for each patient. As shown in Table \ref{tab2}, causal estimates are presented for each QoL domain score (i.e., physical health and psychological). As shown in this table, the ATE estimates for all three time-points were positive, meaning that the Active group had higher QoL scores in physical health. For example, the estimated effect of being physically active in week 15 was calculated as 10, showing that being active causes a 10-unit increase in perception of the physical health score domain when pregnant women are considered active. Similarly, in week 34 of pregnancy, the ATE value was 4.41, meaning that the active group had a better physical health score by 4 units. In addition, in post-partum, an ATE  of 7.03 is achieved, showing an increase of 3 units in physical health score compared to week 34.

Likewise, in the psychological domain, in all three timepoints, the ATE value was positive, showing that active pregnant women have reported a higher psychological score. For instance, during weeks 15 and 34, the ATE values were 2.81 and 2.67, respectively, meaning that active pregnant women have higher psychological scores during early pregnancy and mid-pregnancy by roughly 2 units. 
Moreover, it is worth mentioning that the ATE values for both physical health and psychological domains have a higher magnitude in the first timepoint (i.e., week 15 of pregnancy) and the last timepoint (i.e., post-partum week 12) compared to week 34. This variation in ATE values is due to the fact that the perception of  QoL is affected strongly by the progression of pregnancy \cite{b7}. Since QoL is likely to be affected by multiple factors, determining individual effects is difficult.

\subsection{Refutation Analysis}\label{SCM}
In this study, four refutation analysis methods were used to validate the estimated results obtained by causal discovery. Table \ref{tab2} summarizes the refutation results. Our estimated causal effects were validated by all four refutation tests. Based on the p-values (which are higher than 0.05), it is evident that there are no significant causal effects in the refutation scenarios. Moreover, we were able to confirm the robustness of our initial estimation by considering the new estimate for the placebo treatment, which was approximately zero, and the new estimates for the other methods remain unchanged.

In summary, the proposed CML showed having more PA during pregnancy and post-partum increases the QoL score in a causal way. Our causal graph incorporated the direct and indirect effects of the PA parameters as well as the medical and context information collected subjectively to minimize the bias of estimates. Moreover, to have a better understanding and meaningful interpretation of QoL scores, we classified QoL scores as (100–95)  Near perfect QoL, (95–85) Very good QoL, (85–70) Good QoL, (70—57.5) Moderately QoL, and (57.5–40) Somewhat bad QoL \cite{b22}. As shown in the result, there is an average 7.3-unit increase in each physical health score from low active to active group, indicating changes in an individual's QoL experience (for example, from Near perfect to Very good). In future work, we extend the CML method to personalized treatment effects and incorporate transportability into the model for the transfer of causal effects to a new population.
\section{CONCLUSION}
This paper explored the causal effect of PA on QoL among pregnant women. Maternal (medical) history and context data, as well as wearable-derived data, were also considered crucial factors in our analysis. In this regard, a CML method was developed, including both causal discovery and causal inference components. Based on our findings, we can make an inference about how the PA affects pregnant women's QoL. Our result indicated that the PA led to an improvement in the QoL during pregnancy and post-partum. As part of our causal graph generation efforts, we incorporated domain expertise. 
% \usepackage{multirow}
\begin{table}[tbp]
\setlength{\tabcolsep}{1.25pt}
\label{tab2}
\caption{Validation result for proposed causal graph}
\begin{tabular}{|c|cccccc|}
\hline
\multirow{3}{*}{Method}                                                        & \multicolumn{3}{c|}{\textbf{Physical Health Score}}                                                                                                                                        & \multicolumn{3}{c|}{\textbf{Psychological Score}}                                                                                                                     \\ \cline{2-7} 
                                                                               & \textbf{\begin{tabular}[c]{@{}c@{}}Estimated \\ effect\end{tabular}} & \textbf{\begin{tabular}[c]{@{}c@{}}New \\ estimated\\  effect\end{tabular}} & \multicolumn{1}{c|}{\textbf{p-value}} & \textbf{\begin{tabular}[c]{@{}c@{}}Estimated \\ effect\end{tabular}} & \textbf{\begin{tabular}[c]{@{}c@{}}New\\  estimated \\ effect\end{tabular}} & \textbf{p-value} \\ \cline{2-7} 
                                                                               & \multicolumn{6}{c|}{\textbf{Week 15}}                                                                                                                                                                                                                                                                                                                              \\ \hline
\textbf{\begin{tabular}[c]{@{}c@{}}Placebo\\  treatment\end{tabular}}          & 10.08                                                                & 0.48                                                                        & \multicolumn{1}{c|}{0.38}             & 2.81                                                                 & 0.18                                                                        & 0.42             \\ \hline
\textbf{\begin{tabular}[c]{@{}c@{}}Add random  cause\end{tabular}}           & 10.08                                                                & 9.88                                                                        & \multicolumn{1}{c|}{0.38}             & 2.81                                                                 & 2.98                                                                        & 0.38             \\ \hline
\textbf{Data subset}                                                           & 10.08                                                                & 9.95                                                                        & \multicolumn{1}{c|}{0.32}             & 2.81                                                                 & 2.40                                                                        & 0.29             \\ \hline
\textbf{\begin{tabular}[c]{@{}c@{}}Unobserved \\ random  cause\end{tabular}} & 10.08                                                                & 10.04                                                                       & \multicolumn{1}{c|}{—}                & 2.81                                                                 & 2.82                                                                        & —                \\ \hline
\textbf{}                                                                      & \multicolumn{6}{c|}{\textbf{Week 34}}                                                                                                                                                                                                                                                                                                                              \\ \hline
\textbf{\begin{tabular}[c]{@{}c@{}}Placebo \\ treatment\end{tabular}}          & 4.41                                                                 & 0.19                                                                        & \multicolumn{1}{c|}{0.36}             & 2.67                                                                 & -0.02                                                                       & 0.48             \\ \hline
\textbf{\begin{tabular}[c]{@{}c@{}}Add random  cause\end{tabular}}           & 4.41                                                                 & 0.41                                                                        & \multicolumn{1}{c|}{0.45}             & 2.67                                                                 & 2.70                                                                        & 0.82             \\ \hline
\textbf{Data subset}                                                           & 4.41                                                                 & 3.90                                                                        & \multicolumn{1}{c|}{0.25}             & 2.67                                                                 & 2.26                                                                        & 0.25             \\ \hline
\textbf{\begin{tabular}[c]{@{}c@{}}Unobserved\\  random  cause\end{tabular}} & 4.41                                                                 & 0.47                                                                        & \multicolumn{1}{c|}{—}                & 2.67                                                                 & 2.65                                                                        & —                \\ \hline
\textbf{}                                                                      & \multicolumn{6}{c|}{\textbf{Post-Partum Week 12}}                                                                                                                                                                                                                                                                                                                  \\ \hline
\textbf{\begin{tabular}[c]{@{}c@{}}Placebo \\ treatment\end{tabular}}          & 7.03                                                                 & 0.01                                                                        & \multicolumn{1}{c|}{0.32}             & 5.04                                                                 & 0.40                                                                        & 0.46             \\ \hline
\textbf{\begin{tabular}[c]{@{}c@{}}Add random cause\end{tabular}}           & 7.03                                                                 & 7.02                                                                        & \multicolumn{1}{c|}{0.49}             & 5.04                                                                 & 4.98                                                                        & 0.43             \\ \hline
\textbf{Data subset}                                                           & 7.03                                                                 & 7.36                                                                        & \multicolumn{1}{c|}{0.47}             & 5.04                                                                 & 5.53                                                                        & 0.19             \\ \hline
\textbf{\begin{tabular}[c]{@{}c@{}}Unobserved \\ random  cause\end{tabular}} & 7.03                                                                 & 7.06                                                                        & \multicolumn{1}{c|}{—}                & 5.04                                                                 & 5.82                                                                        & —                \\ \hline
\end{tabular}
\end{table}
% \begin{table}[htbp]
% \caption{Table Type Styles}
% \begin{center}
% \begin{tabular}{|c|c|c|c|}
% \hline
% \textbf{Method}&\multicolumn{3}{|c|}{\textbf{Week 12}} \\
% \cline{2-4} 
% \textbf{Name} & \textbf{\textit{
% Estimated Effect}}& \textbf{\textit{New Estimated Effect}}& \textbf{\textit{p-value}} \\
% \hline

% Placebo treatment & More & &  \\
% \hline
% Add Random Cause& More & &  \\
% \hline
% Data Subset& More & &  \\
% \hline
% Unobserved common cause& More & &  \\
% \hline
% \end{tabular}
% \label{tab1}
% \end{center}
% \end{table}

\vspace{12pt}

\end{document}